\documentclass[runningheads]{llncs}
\usepackage[T1]{fontenc}
\usepackage{multirow}
\usepackage{graphicx}
\usepackage{booktabs}
\usepackage{amsmath}
\usepackage{amssymb}
\usepackage{adjustbox}
\usepackage{subcaption}
\usepackage{pgfplots}
\pgfplotsset{compat=1.18}
\usepackage{multirow}

\makeatletter
\newcommand{\etal}{\emph{et~al}\@ifnextchar.{}{.\ }}
\makeatother

\usepackage[nolist,nohyperlinks]{acronym}
\begin{acronym}
\acro{VAE}{variational autoencoder}
\acro{DDPM}{Denoising Diffusion Probabilistic Model}
\acro{GAN}{generative adversarial network}
\acro{CNN}{convolutional neural network}
\acro{ViT}{vision transformer}
\acro{FPN}{feature pyramid network}
\acro{MRE}{mean radial error}
\acro{SDR}{successful detection rate}
\acro{STD}{standard deviation}
\acro{GT}{ground truth}
\end{acronym}

\usepackage[pagebackref=true,breaklinks=true,colorlinks,bookmarks=false]{hyperref}
\usepackage{cleveref}

\begin{document}
    \title{Optimising for the Unknown: Domain Alignment for Cephalometric Landmark Detection}

%

    \titlerunning{Optimising for the Unknown}
%
    \author{Julian Wyatt\thanks{Corresponding Author} \and Irina Voiculescu}
    \institute{Department of Computer Science, University of Oxford\\
    \email{\{name\}.\{surname\}@cs.ox.ac.uk}}
    \authorrunning{Julian Wyatt \and Irina Voiculescu}

%
    \maketitle              
    \begin{abstract}
        Cephalometric Landmark Detection is the process of identifying key areas for cephalometry. Each landmark is
        a single \acf{GT} point labelled by a clinician. A machine
        learning model predicts the probability locus of a landmark 
        represented by a heatmap. This work,  for the 2024 CL-Detection  MICCAI Challenge, proposes a domain alignment strategy with a regional facial extraction module and an X-ray artefact augmentation procedure. The challenge ranks our method's results as the best in \acf{MRE} of $1.186mm$ and third in the $2mm$ \acf{SDR} of $82.04\%$ on the online validation leaderboard. The code is available at \url{https://github.com/Julian-Wyatt/OptimisingfortheUnknown}.



        \keywords{Landmark Detection  \and ConvNeXt V2 \and RCNN.}
    \end{abstract}

    \section{Introduction}




    Cephalometric Landmark Detection is the task for predicting anatomical keypoints: a single pixel or a neighbourhood of pixels, in an orthodontic X-ray.
    The region of pixels is defined by a clinical orthodontist and is used in cephalometric analysis. The analysis evaluates the spatial relationships between internal facial structures, such as the teeth, jaw, spine and soft tissue leading to craniofacial condition diagnosis~\cite{proffit2006contemporary}. It is then  used for orthognathic treatment, maxillofacial surgery~\cite{proffit2006contemporary,lake1981surgical}, sleep apnea detection and treatment~\cite{deberry1988cephalometric} and other  planning.
    
    Conventional landmark detection approaches rely on manual annotation. This is both time-consuming and prone to large inter-annotator variability. Therefore the automation of landmark detection is key to reducing the annotation time whilst maintaining the clinical quality accuracy. Typically landmarks are utilised in explainable clinical diagnosis such as determining angles~\cite{mccouat2021automatically} and therefore influence their placement. The difficulty in landmark detection arises from understanding local and global semantic facial structure, whilst ensuring that the prediction is precise and accurate within a two millimetre margin. Also, no two images are alike: one may have implants or fillings which occlude the landmark, while a second may have their head $2^\circ$~off-centre causing 3D to 2D projection artefacting.
    Consequently, multi-scale feature self-attention mechanisms are incredibly powerful for understanding the local and global image features, but come with a large caveat. The high memory usage in self-attention makes it difficult to deploy these tools for clinical practise; keeping the accuracy high whilst minimising computation is paramount for real-time clinical deployment.
    
    Furthermore, the imaging data is diverse in resolution and aspect ratio. This can lead a model to underperform even with optimal augmentation strategies. Additionally, when considering out-of-domain imaging protocols, the model is unlikely to perform well if the protocol has not been seen before due to domain shift. Therefore out of distribution~\cite{wyatt2022anoddpm} and domain alignment~\cite{bian2020uncertainty} detection and mitigation strategies can help generalise model performance.

    To overcome these difficulties, we propose a two-step extraction-prediction procedure. First, we utilise a Faster-RCNN cephalometric regional extraction module to predict the region of the image related to the landmarks used for cephalometry, ensuring high performance for out-of-domain imagery. Then, to predict landmark heatmaps, we efficiently capture multi-scale features with a custom ConvNeXt V2 encoder backbone and a lightweight MLP feature pyramid decoder. Lastly, to optimise performance on unseen data, we ensemble the predicted coordinates from 4 cross validated feature pyramid networks.


    Recent developments of anatomical landmark detection employ variants of \acp{CNN} and \acp{ViT} to predict a probability heatmap of likely landmark locations. The highest performing hierarchy of models utilise \acp{FPN}~\cite{lin2017feature} which aggregate multi-resolution features~\cite{ao2023feature}. This has been advanced with bespoke attentive modules such as Zhong \etal and Ye \etal\cite{zhong2019attention,ye2023uncertainty} who attend to multi-level features through the encoding process.

    Most commonly, the heatmap is generated by optimising the negative log likelihood~\cite{mccouat2022contour}, while alternative approaches consider predicting displacement vectors via coordinate regression~\cite{wu2019facial}, or pixel regression using offset maps~\cite{chen2019cephalometric} to improve the overall radial error.




    \section{Methodology}
    \begin{figure}[ht]
        \centering
        \includegraphics[width=.75\linewidth]{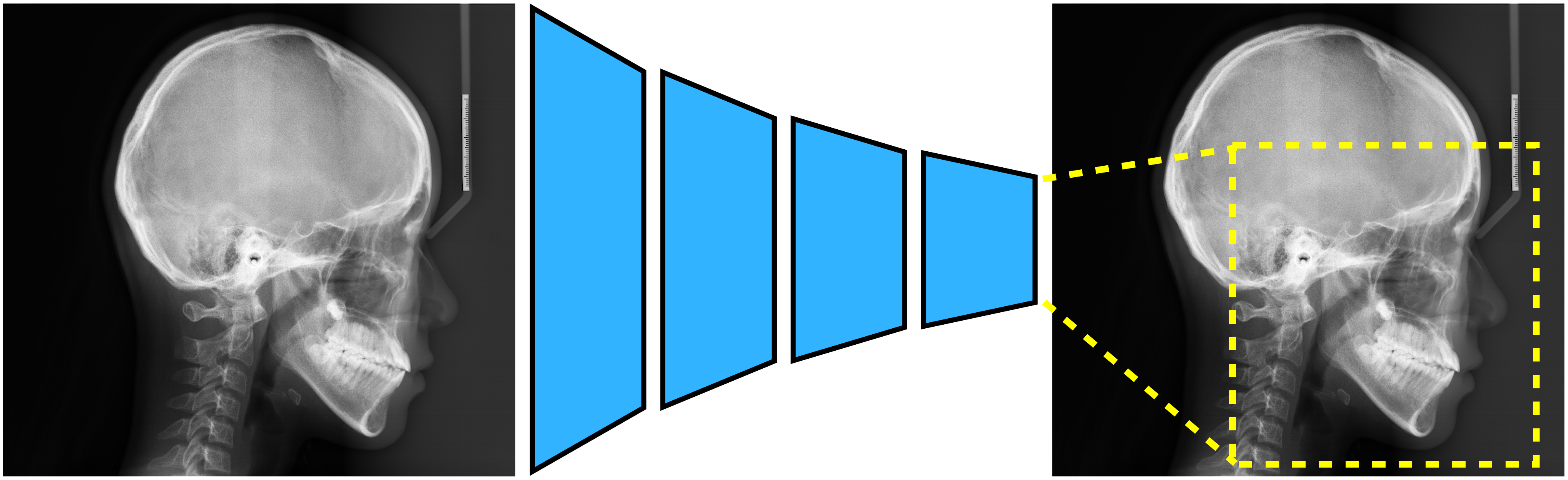}
        \caption{Abstraction of the RCNN regional face extraction module.}
        \label{fig:RCNN}
    \end{figure}
    \subsection{RCNN regional facial extraction}



    An optimal landmark detection model for use in clinical practises must generalise between not only the demographics of patients but also the imaging preprocessing and protocols. To bridge this gap we develop an RCNN regional facial extraction module. This extracts the relevant region of the face where the landmarks are used for cephalometry. 
    The Faster-RCNN model\footnote{\url{https://pytorch.org/vision/stable/models/faster_rcnn.html}}~\cite{ren2016faster} utilises a MobileNet V3 small encoder backbone pretrained on imagenet 1K. Configured with larger anchor sizes of (128, 256, 320, 512), aspect ratios of (0.5, 0.75, 1.0, 1.25, 1.5, 1.75) and the default RoI pooling configurations. The model is then optimised based on the ground truth bounding boxes with the original formulation~\cite{ren2016faster} by jointly optimising the region classification loss and the bounding box regression loss. To define the ground truth, we generate a bounding box offset using the outermost landmarks using an offset padding value of 32 pixels. This value is ablated in \cref{fig:RCNN_ablation}.

    \subsection{Landmark heatmap regression}
   
    \begin{figure}[htbp]
        \centering
        \includegraphics[width=\linewidth]{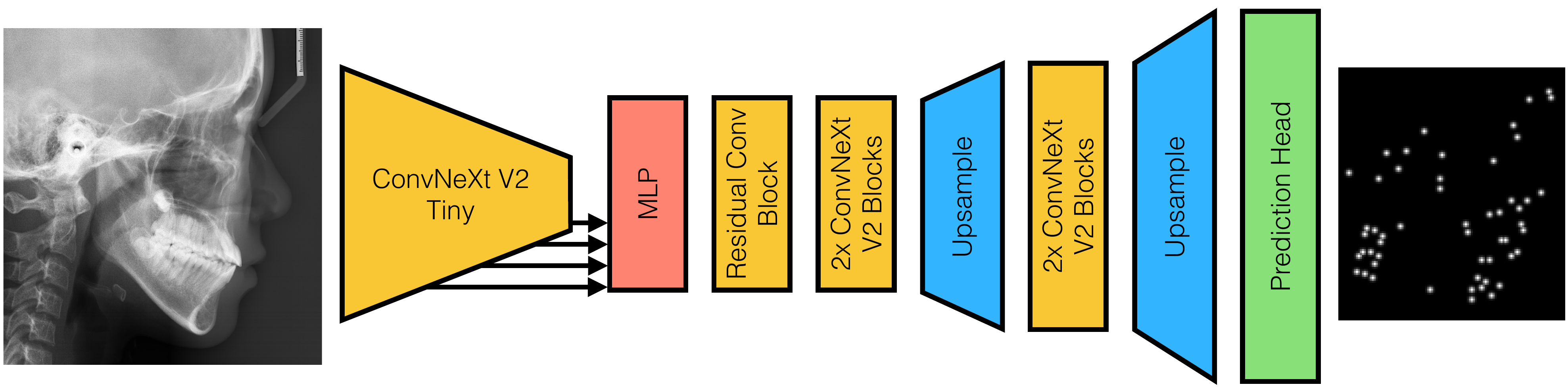}
        \caption{ConvNeXt V2 MLP feature pyramid prediction architecture.}
        \label{fig:pred}
    \end{figure}

    With the recently cropped query image, we predict the final coordinates with heatmap regression. This optimises an $\mathbb{R}^{B\times L \times H \times W}$ tensor, with batch size $B$, total landmarks $L$, height $H$ and width $W$. Larger values within the heatmap indicate a higher probability of proximity to landmark positions.
    
    The loss objective is to minimise the cross entropy between a binary ground truth landmark image $y$ and the model's prediction passed through a 2D softmax activation function $\sigma(\hat{y})$ defined as: $\mathcal{L}=-y\cdot \log (\sigma (\hat{y}))$. The binary ground truth is defined similarly to McCouat \etal~\cite{mccouat2022contour} where each landmark's heatmap contains a 1 at the pixel corresponding to the $x,y$ coordinate of the landmark's label and a 0 elsewhere. We found that the model performed slightly better when Gaussian blurring the ground truth $y$ with standard deviation of 1.


    To modernise the standard ResNet~\cite{he2016deep} U-Net~\cite{ronneberger2015u} configuration, we develop a pretrained ConvNeXt V2~\cite{woo2023convnext} encoder backbone. ConvNeXt is proven to give higher  per-FLOP performance than popular attention mechanisms~\cite{woo2023convnext}, although this is in the image classification domain. To keep this parameter footprint near ResNet34, we explore the self-supervised pretrained nano and tiny variants. Then, to maintain the high model efficacy and influenced from Segformer by Xie \etal~\cite{xie2021segformer} we utilise an MLP feature pyramid with subsequent learned upsampling layers to supersample the feature pyramid to it's original resolution. The learned upsampling layers first convolve the features with a residual convolutional block for channel consistency, followed by 2 ConvNeXt blocks, and a learned convolutional upsampling layer, repeated twice. The final prediction head convolves the final features down to one channel per landmark with preceding group normalisation and the GELU activation function. 
    Furthermore, we set a stochastic drop path~\cite{huang2016deep} rate of 0.375 for the encoder and 0.275 for the up blocks, with 0.2 2D dropout probability for the residual convolution block. 
    

    We mitigate the effect of overfitting, while ensuring high quality predictions by ensembling four models from different cross validated folds. Each model predicts a heatmap for a test image, where the hottest 20 points are averaged for a given model's final landmark prediction. The final coordinates from each model are then further averaged to reach the final ensembled prediction.

    \section{Experiments}

    \subsection{Dataset}
    The dataset for the CL-Detection 2024 challenge comprises approximately 700 dental X-ray lateral images from four different centres.
    This dataset is organised into four distinct subsets: training, validation, test, and independent test sets.
    The training, validation, and test sets all derive from the ISBI challenge dataset~\cite{wang2016benchmark}, the PKU cephalogram dataset~\cite{zeng2021cascaded}, and the Shenzhen University General Hospital dataset~\cite{chen2019cephalometric}. The training set includes 396 images, the validation set has 50 images, and the test set consists of 150 images.
    The independent test set features 100 images obtained from a private medical centre, offering a unique evaluation benchmark.
    Dental professionals have annotated 53 landmarks for each image,
    including 13 soft tissue-related landmarks, 6 tooth-related landmarks, 19 skull-related landmarks, 13 cervical spine-related landmarks and 2 calibration ruler landmarks.


    Our approach to the landmark detection task involves cropping and downsizing the images with the RCNN detection module, such that the height is 800~pixels and the width maintains the predicted box's aspect ratio. Although this limited the batch size to 1, it ensured the predicted coordinates were accurately mapped back to the original resolution for challenge submission.
    


    We followed the evaluation prescribed in the challenge, measuring the \acf{MRE} and the $2mm$ \acf{SDR}.

    \subsection{Implementation details}

    We trained the 4 cross validated models over 75 epochs on a single Linux node with a Tesla V100-SXM2-32GB-LS GPU that held 35\% memory usage. Each model took approximately 4 hours to train. The network is optimised using the AdamW optimiser~\cite{loshchilov2017decoupled} with learning rate $2\times10^{-4}$ and weight decay $0.05$. To improve training times and reduce model variability, we schedule batch accumulation such that at epochs 0, 4 and 8, the batches are accumulated every 32, 16 and 8 steps respectively. The learning rate decayed by a factor of 0.25 at epoch 35 and 45 to reduce overfitting and improve the local minima. The model checkpoint is selected as the best performing epoch's \ac{MRE} on the validation set. Training halts when this has not improved for 10 epochs.

    \subsection{Augmentations and X-ray artefact simulation}

    To further improve generalisability, we swept over popular augmentation techniques, and also propose a new method. The conventional augmentations are primarily implemented using the \texttt{imgaug} library. These include: a rotation of $\pm5^{\circ}$, translations of $\pm 30$ horizontal pixels and $\pm 20$ vertical pixels, scale adjustments between $1\pm0.125$, value multiplication by $1\pm 0.6$, elastic transformations with $\alpha{=}400$ and $\sigma{=}30$, a single cutout with random size between 0.04 and 0.3 of the image size, a random gamma contrast adjustment between 0.3 and 2, colour inversion rate of 0.1, a blur rate of 0.1 and a skewed scale rate of 0.3 where scaling ignores aspect ratio. Lastly, for our novel augmentation method, we seek to introduce X-ray artefacts to improve performance on unseen images. We either add noise sampled from $\mathcal{N}(0,15)$ or multiply by a factor sampled $\mathcal{U}(0.5,1.5)$ to a vertical or horizontal region of the image with size chosen from [25, 50, 75, 100, 125]. An example of the artefacting we aim to reproduce can be seen over the cranium in \cref{fig:RCNN} and most prominently in validation image {\tt 507.bmp}.

    \begin{figure}[hbtp]
    \centering
    \begin{subfigure}{0.32\linewidth}
        \begin{tikzpicture}
        \begin{axis}[
            ylabel={\scriptsize MRE},
            ylabel style={yshift=-2pt},
            width=4.3cm,
            height=3.5cm,
            grid=both, 
            major grid style={line width=0.2pt,draw=gray!50},
            minor grid style={line width=0.1pt,draw=gray!20}, 
            xtick={32,96,160, 200},      
            xticklabels={32,96,\textbf{y},\textbf{z}},   
            y tick style={left}, 
            every axis plot/.append style={thick}, 
            tick label style={font=\footnotesize}, 
        ]
        \addplot table [x=padding, y=MRE_subset, col sep=comma] {csvs/preprocessing_padding_ablation.csv};
        \end{axis}
        \end{tikzpicture}
        \subcaption{\scriptsize RCNN Padding}
        \label{fig:RCNN_ablation}
    \end{subfigure}
    \begin{subfigure}{0.33\linewidth}
        \begin{tikzpicture}
        \begin{axis}[
            ylabel={\empty},
            width=4.3cm,
            height=3.5cm,
            grid=both, 
            major grid style={line width=0.2pt,draw=gray!50}, 
            minor grid style={line width=0.1pt,draw=gray!20}, 
            axis line style={-}, 
            xtick={0,0.1,0.2,0.3,0.4,0.5,0.6,0.7,0.8,0.9,1},      
            xticklabels={0, , .2, , .4, , .6, , .8, , 1},   
            y tick style={left}, 
            every axis plot/.append style={thick}, 
            tick label style={font=\small}, 
        ]
        \addplot table [x=SIMULATE_XRAY_ARTEFACTS_RATE, y=MRE, col sep=comma] {csvs/xray_artefact_simulation.csv};
        \end{axis}
        \end{tikzpicture}
        \subcaption{\scriptsize Artefact Augmentation Rate}
        \label{fig:aug_ablation}
        
    \end{subfigure}
    \begin{subfigure}{0.32\linewidth}
        \begin{tikzpicture}
        \begin{axis}[
            ylabel={\empty},
            ylabel style={yshift=-10pt},
            width=4.3cm,
            height=3.5cm,
            grid=both, 
            major grid style={line width=0.2pt,draw=gray!50}, 
            minor grid style={line width=0.1pt,draw=gray!20}, 
            axis line style={-}, 
            xtick={0, 5,10,15,20, 25},
            xticklabels={0, 5,10,15,20, 25},
            y tick style={left}, 
            every axis plot/.append style={thick}, 
            tick label style={font=\small}, 
        ]
        \addplot table [x=TOP_K_HOTTEST_POINTS, y=TEST_MRE, col sep=comma] {csvs/top_k.csv};
        \end{axis}
        \end{tikzpicture}
        \subcaption{\scriptsize Top K hottest points}
        \label{fig:top_k_ablation}

    \end{subfigure}
    \caption{Ablation studies trained with the nano encoder. \textbf{(a)} Impact of RCNN padding around landmarks for bounding box generation during training from 16-128 pixels. Models are trained on patients 1-400 and evaluated on 401-600. Here, y is a deterministic image pad and crop method and z is a standard pad to aspect ratio and resize. \textbf{(b)} Effect of varying the frequency of X-ray artefact simulation on model performance (Fold~1). \textbf{(c)} Influence of the number of hottest heatmap values averaged to determine the final coordinates (Fold 1).}
    \label{fig:ablation}
    \end{figure}
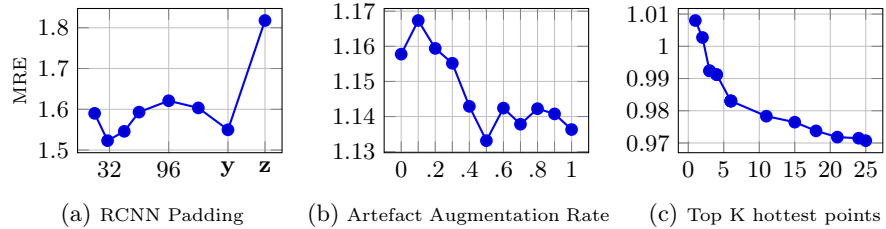

    \section{Results and discussion}

    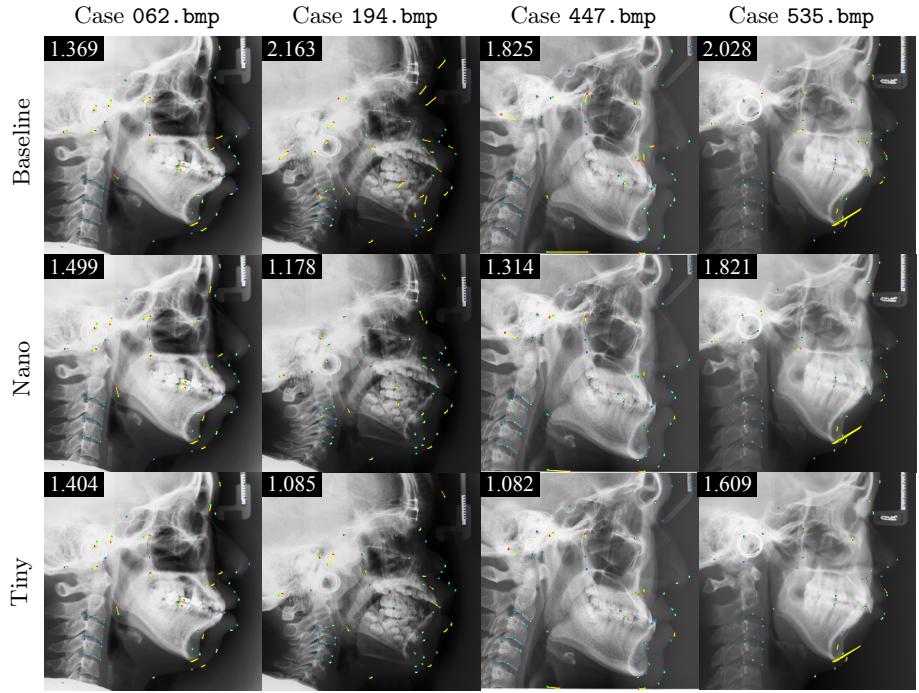
\begin{figure}[t]

      \centering
      \vspace{-0.2cm}
      \begin{tikzpicture}
    
            \node[inner sep=0] (image1) at (-1,0) {\includegraphics[width=0.95\linewidth]{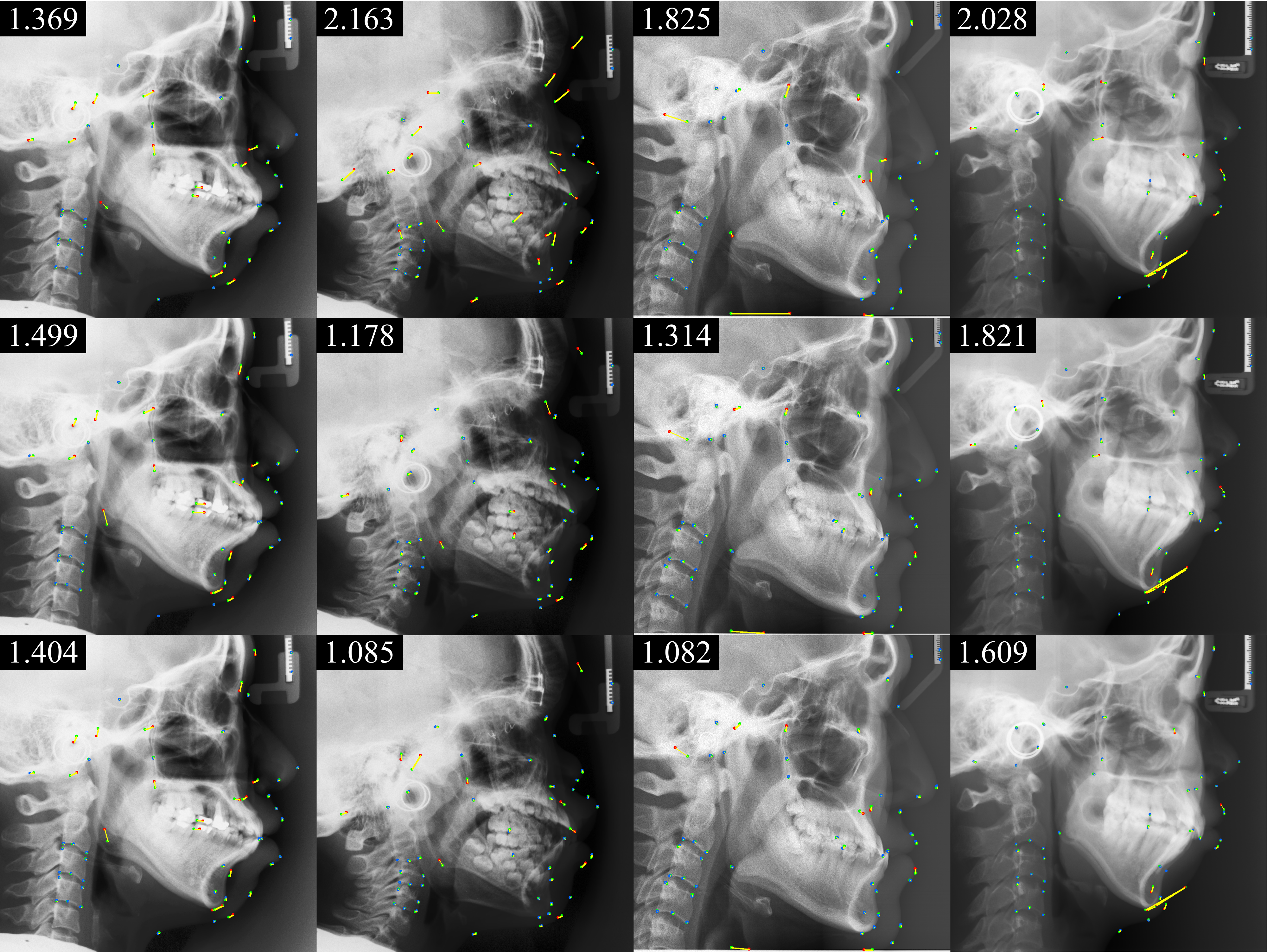}};
    
             \node[rotate=90, text width=1.3cm] at (-7.1,3)  {\small Baseline};
    
            \node[rotate=90, text width=1.3cm] at (-7.1,0.2)  {\small Nano}; 
            \node[rotate=90, text width=1.3cm] at (-7.1,-2.6)  {\small Tiny};
    
            \node[minimum height=0.5cm,text width=2cm] at (-5.4,4.6)  {\small Case {\tt 062.bmp}};
            \node[minimum height=0.5cm,text width=2cm] at (-2.5,4.6)  {\small Case {\tt 194.bmp}};
            \node[minimum height=0.5cm,text width=2cm] at (0.4,4.6)  {\small Case {\tt 447.bmp}};
            \node[minimum height=0.5cm,text width=2cm] at (3.3,4.6)  {\small Case {\tt 535.bmp}};
        \end{tikzpicture}
    
      \caption{Qualitative results for four cases: the worst overall SDR for the tiny encoder ({\tt 062.bmp}), the worst for the baseline ({\tt 194.bmp}), and two hand-picked failure cases ({\tt 447.bmp} \& {\tt 535.bmp}). Blue dots represent predictions within $2mm$, red dots are outside $2mm$, and green dots are ground truth, with yellow lines connecting predictions to ground truth. The top-left number indicates the MRE for each image, and all images are cropped to square to highlight landmarks rather than by the RCNN module.}
      \label{fig:qualitative}
    \end{figure}






    Our experiments were conducted by first analysing the online validation performance of our baseline model~\cite{mccouat2022contour} using a deterministic preprocessing strategy (y in \cref{fig:RCNN_ablation}). We then subsequently used the best landmarks from this as local labels to maximise the amount of available data. This model scored an \ac{MRE} of $1.218mm$, and an \ac{SDR} of 82.03\%. Following this we defined 4 cross validation folds for later use in ensembling. 
    
    Throughout the remainder of development, we performed several ablation studies and have chosen a notable few as seen in \cref{fig:ablation}. Firstly, we evaluated the performance of domain translation by training the landmark detection model on the first data subset (patients 1-400) and evaluating on the remaining images. Unsurprisingly, utilising the RCNN feature extraction module over just padding and resizing (z in \cref{fig:RCNN_ablation}) helps localise the face and improve performance to unseen images from different domains. 
    The second ablation shown in \cref{fig:aug_ablation} shows the effect of how the frequency of our novel augmentation strategy influences the best \ac{MRE}: higher values improve the final \ac{MRE}. From our augmentation sweep, we balance \ac{SDR} and \ac{MRE} performance by selecting 0.9 as a rate of including this augmentation method. Lastly, we experiment with how many of the hottest points in the heatmap are averaged to determine the final coordinates for a given landmark. When downsampling images, the ground truth landmark coordinates may reside between pixels and is discretised for heatmap regression. Therefore by averaging multiple points, the pixelwise radial error improves. We chose 20 points as this balanced both \ac{MRE} and \ac{SDR}. Moreover, we ablated methods to convert the pretrained ConvNeXt model from 3 channel input to a single channel.
    Out of repeating the input data, convolving, and summing the original initial weights~\cite{mccouat2022contour}, the latter performed the best.

    To best highlight failure cases, we show 4 patients in \cref{fig:qualitative}. The images are selected as the worst performing SDR for the tiny encoder ({\tt 062.bmp}), worst SDR for the baseline ({\tt 194.bmp}) and two hand-picked failure cases from the other two domains ({\tt 447.bmp} \& {\tt 535.bmp}). All images are taken as final ensembled predictions and will therefore have been trained on 3/4 submodels as we did not have validation labels. Despite this, the performance is clearly still suboptimal.
    
    Case {\tt 062.bmp} struggles due to occlusions around the teeth, incorrect labelling around the bone chin landmarks and lastly, around the Gonion landmark (end of the jaw), the ground truth landmark is not near a meaningful boundary. This is likely a result of averaging two clinical labels -- a process abstracted away from challenge participants. Case {\tt 194.bmp} is far out of distribution, with difficult boundary contrast, unerupted teeth and small features. The high performance on the tiny and nano models is likely a result of overfitting to this example. Case {\tt 447.bmp} is an edge case for the RCNN feature extraction module as the bottom two landmarks are too close to the boundary. This therefore is affected by not only the preprocessing but also the affine augmentations. Lastly, case {\tt 535.bmp} is an extreme example of mislabelled data around the chin and almost every other landmark is correctly detected for the tiny encoder. 

    \begin{table}[t]
    \caption{Four fold cross validation performance with ensembling performance across every training image.}
    \label{tab:cross_val_results}
    \centering
    \begin{tabular}{c|cc|cc|cc|cc|cc}
    \toprule
    \multirow{2}{*}{Method} &
      \multicolumn{2}{c|}{Fold 1} &
      \multicolumn{2}{c|}{Fold 2} &
      \multicolumn{2}{c|}{Fold 3} &
      \multicolumn{2}{c|}{Fold 4} &
      \multicolumn{2}{c}{Ensemble} \\ 
     &
      \multicolumn{1}{c}{MRE} &
      SDR &
      \multicolumn{1}{c}{MRE} &
      SDR &
      \multicolumn{1}{c}{MRE} &
      SDR &
      \multicolumn{1}{c}{MRE} &
      SDR &
      \multicolumn{1}{c}{MRE} &
      SDR \\ \hline
    
    
    
    Baseline~\cite{mccouat2022contour} (CHH)  &  1.20 & 83.12 & 1.21 & 82.44 &  1.29 & 81.56 & 1.25 & 82.19 &  1.22 & 82.42 \\ 
    Nano Encoder & 1.16 & 84.13 & 1.16 & 83.87 & 1.24 & 82.78 & 1.20 & 83.40  & 0.97 & 87.85 \\ 
    Tiny Encoder &  1.18 & 84.13 &  1.16 & 83.68 & 1.24 & 82.49 & 1.20 & 83.20 & 0.93 & 88.61  \\
    
    \bottomrule
    \end{tabular}
    \end{table} 

    Highlighted in \cref{tab:cross_val_results}, each model performs similarly on the individual cross validation folds. However, the ensembled performance, measuring the performance across every real labelled image, increases with model size. This is potentially a result of overfitting to the individual folds which we mitigate with higher weight decays, early stopping and selecting best epoch validation performance per fold as the ensembled checkpoint.

    We chose the tiny encoder over the nano encoder as our final model because the online validation MRE improved from 1.20 to 1.19, while the SDR improved by $2\%$. Despite the runtime only increasing by 2 seconds and slight memory increases, the additional performance was sufficient for the increase in compute. 

    \subsection{Limitations \& future work}


    While generating strong performance, there are still some limitations with the model. Primarily, the scale of the ConvNeXt models leads to heavy overfitting shown as a degradation in validation performance over time. Combating this with stochastic drop path~\cite{huang2016deep}, also has flaws as it results in incredibly inconsistent validation performance: a poorly chosen seed can heavily influence the \ac{MRE}.

    In the future, we would like to explore the potential in further refining the RCNN module to predict bounding boxes for proximally similar landmarks, for example, leading to separate images for the spine, chin, ear and eye.



    
    \begin{table}[t]
        \caption{Quantitative evaluation results for final submitted model. Hidden and independent tests resulted in 26.3 seconds runtime with area under GPU-memory time curve of 13488 MiBs}
        \label{tab:final-results}
        \centering
        \renewcommand{\arraystretch}{1.25}
        \begin{adjustbox}{width=\textwidth}
            \begin{tabular}{l|cc|cc|cc}
                \toprule
                \multirow{2}{*}{Method} & \multicolumn{2}{c|}{Online Validation} & \multicolumn{2}{c|}{Hidden Test}  & \multicolumn{2}{c}{Independent Test}     \\
                \cline{2-7}
                & MRE (mm)          & SDR@2mm (\%)       & MRE (mm)          & SDR@2mm (\%)       & MRE (mm)          & SDR@2mm (\%)       \\ \hline
                Tiny Encoder &  1.186 & 82.04                 &  1.24 & 81.9 &  3.27 & 57.9                                       \\
                \bottomrule
            \end{tabular}
        \end{adjustbox}
    \end{table}








    \section{Conclusion}
    In this work, we primarily address the domain gap between cephalometric imaging from diverse sources. By integrating an RCNN regional extraction preprocessing stage and a novel X-ray artefacting augmentation strategy, we enhance the generalisability to unseen data from alternative domains. The approach not only improves the robustness to artefacting, crucial for clinical deployment with similar real-world issues, but also keeps the runtime and memory footprint low for practical clinical deployment.

    \subsubsection{Acknowledgements.} The authors of this paper declare that the landmark detection method they implemented for participation in the CL-Detection 2024 challenge has not used any additional datasets other than those provided by the organisers.
    The proposed solution is fully automatic without any manual intervention.
    We thank all the data owners for making the X-ray images and CT scans publicly available and Codabench~\cite{xu2022codabench} for hosting the challenge platform.
    This work was funded by an EPSRC
    Research Scholarship.
    The authors would like to acknowledge the use of
    the University of Oxford Advanced Research Computing (ARC) in carrying out this work.\footnote{\url{http://dx.doi.org/10.5281/zenodo.22558}}

%
%
%
    \bibliographystyle{splncs04}
    \bibliography{refs}

\end{document}